\documentclass[runningheads]{llncs}
\usepackage[T1]{fontenc}
\usepackage{graphicx}
\usepackage{booktabs}
\usepackage[misc]{ifsym}
\newcommand{\corr}{(\Letter)}

\begin{document}

\title{On the Instance Hardness as a Decision Criterion in TinyML Systems}
\titlerunning{On the Instance Hardness in TinyML Systems}

\author{Tobiasz Puslecki \corr \and Krzysztof Walkowiak}
\authorrunning{T. Puslecki and K. Walkowiak}
\institute{Department of Systems and Computer Networks, \\Wroclaw University of Science and Technology\\ \email{tobiasz.puslecki@pwr.edu.pl}}

\maketitle

\begin{abstract}
TinyML includes the implementation of machine learning on devices with limited memory and computing resources. With the development of technology, AI systems continue to scale in terms of size and computational requirements. This forces researchers to adapt methods to be environmentally sustainable by designing techniques for reducing computational costs and energy consumption in inferring AI models, even in small devices. In this work, we present preliminary findings on a novel application of the tree depth prune instance hardness method to the TinyML system. The results indicate that threshold control can change energy consumption with limited classification quality changes. This method allows us to adjust classification accuracy, thereby influencing computational complexity and energy consumption for inference. We present a work in progress with initial results as a proof of concept.
\end{abstract}

\section{Introduction}\label{sec:introduction}
TinyML includes the implementation of machine learning models on devices with limited memory and computing resources~\cite{tdp_warden}. These models must be lightweight and powerful to meet strict real-time constraints~\cite{tdp_tinySurvey}. In many TinyML applications, power consumption is a critical limitation, so processing data in an energy-efficient way while maintaining the accuracy of the model is the main challenge~\cite{tdp_tinySurvex}. With the development of technology, artificial intelligence systems continue to scale in terms of size and computational requirements. This forces researchers to adapt methods to be environmentally sustainable by designing techniques for reducing computational costs and energy consumption in inferring AI models, even in small devices. Another problem with TinyML systems is the changing context in which these devices operate. Data distribution, battery charge status, data intensity in the datastream, or the availability of energy from external energy sources (e.g., photovoltaic panels) may change.

Instance hardness (IH) is a set of measures that indicates the degree of difficulty in classifying an instance~\cite{tdp_LorenaTrusting}. In general, the higher the hardness value, the more difficult it is for the model to trust the correct label of that instance. It is worth noting that samples subjected to inference may vary in IH. Consequently, some of the easier samples can be processed by a less complex model, while the rest can be handled by a much more complex one (locally or in the cloud, depending on the use case). This allows, through a trade-off between quality and complexity, for energy resources not to be wasted. Due to the nature of TinyML systems, labels for continuous evaluation cannot be obtained immediately or at all. This necessitates the use of unsupervised methods to calculate IH. One method for measuring IH is tree depth (TD)~\cite{tdp_Smith2014}, i.e., the depth of the leaf node that classifies a given instance in the decision tree, normalized by the maximum depth of the tree. There are two versions of this metric using pruned (TDP) or unpruned (TDU) decision trees. Instances that are harder to classify are typically placed at deeper levels of the tree and have higher TD values (they are close to the decision boundary). TD could be used to decide whether to use a lightweight classifier or a heavier ensemble~\cite{tdp_paper0,tdp_paper1}. TD can also be applied in concepts similar to big/LITTLE DNNs~\cite{tdp_bigLITTLE} (a large DNN is used when the output of a small DNN is estimated to be inaccurate), helping to decide when to use a large DNN.\newline
\indent The need to investigate new energy-saving methods necessitates exploring approaches other than deep neural networks (DNNs). Ensemble methods may serve as an alternative to deep learning methods~\cite{tdp_ensemble}. In ensemble methods, multiple weak models are trained offline, and at runtime, the system selects the appropriate models based on the conditions. For example, a smaller ensemble can be used when currently available resources are limited, otherwise a larger and more complex ensemble is used. A key feature in the construction of the ensemble pool is diversity. It can reduce the risk of overfitting and improve the ability to generalize. Most often, it is achieved by choosing classifiers from different model families or different sets of hyperparameters. Different base models may be appropriate in different regions of the local feature space for a given sample. For this reason, a convenient approach is to select the most competent models from a pool to form a predictive ensemble. This approach is commonly known as Dynamic Ensemble Selection (DES). Unlike static selection, where the choice of classifiers is made during the training phase, dynamic selection (DS) is performed for each new test sample during the classification phase. DES systems can be an alternative for improving the overall accuracy of the system compared to monolithic classifiers, such as DNNs.~\cite{tdp_2a}. 
The most popular DES methods are based on neighborhood (e.g., KNORA-U/E) or clustering (e.g., DES-Clustering). Dynamic neighborhood-based selection methods are not a good choice for resource-constrained systems—it is necessary to keep the validation set in memory to evaluate regions of competence. Although a naive version of the \textit{k}NN algorithm is easy to implement by calculating the distance from a test example to all stored examples, it is computationally intensive for large training sets. Therefore, DS methods based on clustering are of interest for TinyML systems. In this case, only \textit{N} centroids are stored in memory, along with pools of base classifiers assigned to them. DS methods based on clustering, unlike neighborhood-based methods, are heavier during the training phase but lighter during the inference phase~\cite{tdp_item24}.

In~\cite{tdp_paper0}, the authors investigate why dynamic classifier selection methods achieve better results despite using a similar neighborhood as K-NN. They confirm the hypothesis that DES performs better on difficult samples—that is, points near decision boundaries with high IH. In~\cite{tdp_paper1}, the authors expand on this concept and use IH as a decision criterion in dynamic classifier ensembles (KNORA-U and KNORA-E). The authors propose classifying easy samples with a simple single classifier and directing only difficult ones to a more complex system combining dynamic feature selection and dynamic classifier selection. The goal is to reduce computational complexity without a significant loss of classification accuracy. They use kDN, based on the kNN method, as a measure of IH in the process of estimating the local competence of the base classifiers. Both papers address classification using dynamic ensembles and IH measures: the first explains when DES is better than K-NN, and the second shows how to use the same idea of IH to speed up the operation of dynamic ensembles. Importantly, the authors consider a fixed IH threshold but do not consider a scenario in which the threshold could be selected dynamically, depending on the context. Furthermore, both papers use neighborhood-based IH and DES methods, which, as described earlier, are not beneficial for TinyML systems.

Based on the above information, we propose using TDP as an IH metric in combination with the DES-Clustering method, providing a memory- and time-efficient solution tailored to TinyML systems. The main goal of this work is to explore the TDP IH method for the TinyML system. To this end, this paper presents preliminary findings on this novel application of the TDP method.
% Based on the above information, we propose using TDP as an IH metric in combination with the DES-Clustering method as a memory- and time-efficient solution tailored to TinyML systems. In this paper, we present preliminary findings on a novel application of the TDP IH method to the TinyML system. To the best of our knowledge, no one has combined these fields before. We present a work in progress with initial results as a proof of concept.

% The remainder of this article is organized as follows. Section \ref{sec:introduction} contains the introduction and presents the related works. Section \ref{sec:method} describes the method proposed in the article. Section \ref{sec:exp} presents the results of the conducted experiments. Finally, Section~\ref{sec:conclusions} concludes the work.

\section{Method}\label{sec:method}
The main idea behind the presented method is as follows. TDP is used to determine IH. Its normalized leaf depth for a sample serves as a simple measure of IH. Importantly, it is not necessary to use a separate simple classifier. The role of both the simple classifier and the model calculating IH is performed here by the same model, i.e., the pruned decision tree. The IH value lies in the range [0,1], and depending on the set threshold, easy cases ($TDP_{x_i}$ <= threshold — the IH value is less than or equal to the IH threshold) are classified by the decision tree, while difficult cases ($TDP_{x_i}$ > threshold — instances higher than the defined threshold) are passed to the computationally expensive DES-Clustering model.

DES-Clustering is a method that selects an ensemble of classifiers by taking into account the accuracy and diversity of the base classifiers. The \textit{K}-means algorithm is used to define the region of competence. For each cluster, the \textit{N} most accurate classifiers are first selected. Then, the \textit{J} most diverse classifiers from the \textit{N} most accurate classifiers are selected to compose the ensemble~\cite{tdp_1a,tdp_2a,tdp_6a}. The DES-Clustering method works well with TinyML, unlike neighborhood-based methods such as KNORA-U/E~\cite{tdp_item24}.

It is worth noting that, since memory is very limited in TinyML systems, storing depth information for each leaf is expensive. To save memory—equivalent to the number of leaves multiplied by the data type size—the tree-depth proxy may be represented by latency. In TinyML, latency can be obtained from built-in cycle counters or timers and does not require additional computations. Technically, the difference between counter values before and after inference can itself be analyzed, without converting it into specific time units. Thus, leaf depth is analyzed under the assumption that more difficult instances enter deeper into the pruned tree and therefore have longer inference times, because a deeper tree requires a larger number of comparisons along the path from the root to the leaf. Therefore, these two measures—node depth and latency—can be treated as equivalent tools for change detection. This relationship is not perfectly linear because the cost of prediction depends on the average, rather than the maximum, path length, and on the unbalanced structure of the tree.

Minimal cost-complexity pruning is an algorithm used to prune a tree to avoid overfitting. This algorithm is parameterized by the complexity parameter~\cite{tdp_Breiman}. Using the \textit{cost-complexity} parameter (in this research, we use 0.005), the minimal cost-complexity pruning procedure prunes branches with small gains in impurity reduction. Consequently, the higher the density of samples from different classes in a given region, close to the decision boundary, the more conditions need to be checked before a decision is made. Therefore, node depth near the decision boundary, for difficult samples, will be greater.

The threshold works as follows: the higher the threshold, the more instances are classified by the simple TDP tree; the lower the threshold, the more instances are sent to the DES. In other words, when the threshold is 0.00, it means that almost all samples will be sent to the DES, except for those ending at the root of the tree, if any exist. When the threshold is 1.00, it means that all samples will be considered easy and classified by the TDP tree. Intermediate values can represent a compromise between a simple classifier and DES. The IH threshold can be controlled dynamically and contextually, i.e., using a lightweight selector, such as FIS described in ~\cite{tdp_ITEM25}, controlling the accuracy-energy consumption-latency tradeoff. In the context of variable energy levels, the IH threshold can be mapped to the IH threshold, which will have a direct impact on accuracy, energy consumption, and latency.

\section{Experimental evaluation}\label{sec:exp}

\begin{table}[htbp]
\centering
\caption{Results for various methods, datasets, and thresholds, reported as means with standard deviations in parentheses.}
\label{tab:threshold_results}
\resizebox{\textwidth}{!}{%
\begin{tabular}{c|cc|cc|cc}
\hline
\textbf{Threshold}
& \textbf{KNORA-E Accuracy}
& \textbf{KNORA-E Energy Proxy}
& \textbf{KNORA-U Accuracy}
& \textbf{KNORA-U Energy Proxy}
& \textbf{DESC Accuracy}
& \textbf{DESC Energy Proxy} \\
\hline
\multicolumn{7}{c}{\textbf{Vehicle dataset}} \\
\hline
0.0 & 0.671 (0.012) & 0.962 (0.011) & 0.720 (0.014) & 0.962 (0.011) & 0.703 (0.020) & 1.000 (0.000) \\
0.1 & 0.673 (0.010) & 0.844 (0.026) & 0.719 (0.016) & 0.844 (0.026) & 0.708 (0.016) & 1.000 (0.000) \\
0.2 & 0.674 (0.009) & 0.716 (0.035) & 0.715 (0.014) & 0.716 (0.035) & 0.704 (0.025) & 0.979 (0.062) \\
0.3 & 0.681 (0.010) & 0.507 (0.041) & 0.709 (0.016) & 0.507 (0.041) & 0.703 (0.022) & 0.921 (0.098) \\
0.4 & 0.681 (0.010) & 0.507 (0.041) & 0.709 (0.016) & 0.507 (0.041) & 0.687 (0.016) & 0.742 (0.155) \\
0.5 & 0.672 (0.011) & 0.194 (0.025) & 0.680 (0.013) & 0.194 (0.025) & 0.687 (0.022) & 0.605 (0.157) \\
0.6 & 0.648 (0.018) & 0.015 (0.004) & 0.650 (0.017) & 0.015 (0.004) & 0.677 (0.018) & 0.449 (0.195) \\
0.7 & 0.648 (0.018) & 0.015 (0.004) & 0.650 (0.017) & 0.015 (0.004) & 0.674 (0.023) & 0.300 (0.125) \\
0.8 & 0.645 (0.017) & 0.000 (0.000) & 0.645 (0.017) & 0.000 (0.000) & 0.664 (0.023) & 0.162 (0.069) \\
0.9 & 0.645 (0.017) & 0.000 (0.000) & 0.645 (0.017) & 0.000 (0.000) & 0.652 (0.032) & 0.071 (0.039) \\
1.0 & 0.645 (0.017) & 0.000 (0.000) & 0.645 (0.017) & 0.000 (0.000) & 0.650 (0.031) & 0.000 (0.000) \\
\hline
\multicolumn{7}{c}{\textbf{Wine dataset}} \\
\hline
0.0 & 0.951 (0.018) & 0.879 (0.050) & 0.954 (0.042) & 0.879 (0.050) & 0.947 (0.029) & 1.000 (0.000) \\
0.1 & 0.952 (0.018) & 0.347 (0.060) & 0.960 (0.026) & 0.347 (0.060) & 0.952 (0.031) & 1.000 (0.000) \\
0.2 & 0.957 (0.021) & 0.179 (0.051) & 0.962 (0.021) & 0.179 (0.051) & 0.949 (0.034) & 1.000 (0.000) \\
0.3 & 0.965 (0.020) & 0.076 (0.022) & 0.965 (0.024) & 0.076 (0.022) & 0.945 (0.031) & 0.927 (0.147) \\
0.4 & 0.965 (0.020) & 0.076 (0.022) & 0.965 (0.024) & 0.076 (0.022) & 0.947 (0.034) & 0.897 (0.160) \\
0.5 & 0.955 (0.023) & 0.006 (0.006) & 0.955 (0.023) & 0.006 (0.006) & 0.943 (0.044) & 0.861 (0.222) \\
0.6 & 0.954 (0.020) & 0.000 (0.000) & 0.954 (0.020) & 0.000 (0.000) & 0.947 (0.030) & 0.861 (0.222) \\
0.7 & 0.954 (0.020) & 0.000 (0.000) & 0.954 (0.020) & 0.000 (0.000) & 0.928 (0.029) & 0.544 (0.249) \\
0.8 & 0.954 (0.020) & 0.000 (0.000) & 0.954 (0.020) & 0.000 (0.000) & 0.927 (0.032) & 0.536 (0.254) \\
0.9 & 0.954 (0.020) & 0.000 (0.000) & 0.954 (0.020) & 0.000 (0.000) & 0.928 (0.032) & 0.536 (0.254) \\
1.0 & 0.954 (0.020) & 0.000 (0.000) & 0.954 (0.020) & 0.000 (0.000) & 0.858 (0.050) & 0.000 (0.000) \\
\hline
\multicolumn{7}{c}{\textbf{Digits dataset}} \\
\hline
0.0 & 0.874 (0.014) & 0.967 (0.006) & 0.944 (0.007) & 0.967 (0.006) & 0.909 (0.008) & 1.000 (0.000) \\
0.1 & 0.894 (0.011) & 0.380 (0.014) & 0.947 (0.007) & 0.380 (0.014) & 0.906 (0.010) & 1.000 (0.000) \\
0.2 & 0.913 (0.010) & 0.224 (0.014) & 0.951 (0.007) & 0.224 (0.014) & 0.908 (0.011) & 1.000 (0.000) \\
0.3 & 0.928 (0.011) & 0.127 (0.010) & 0.950 (0.006) & 0.127 (0.010) & 0.903 (0.009) & 0.967 (0.059) \\
0.4 & 0.928 (0.011) & 0.127 (0.010) & 0.950 (0.006) & 0.127 (0.010) & 0.893 (0.011) & 0.820 (0.077) \\
0.5 & 0.934 (0.010) & 0.042 (0.007) & 0.940 (0.010) & 0.042 (0.007) & 0.875 (0.012) & 0.670 (0.074) \\
0.6 & 0.933 (0.009) & 0.005 (0.002) & 0.933 (0.010) & 0.005 (0.002) & 0.852 (0.017) & 0.525 (0.140) \\
0.7 & 0.933 (0.009) & 0.005 (0.002) & 0.933 (0.010) & 0.005 (0.002) & 0.826 (0.018) & 0.329 (0.116) \\
0.8 & 0.932 (0.010) & 0.000 (0.000) & 0.932 (0.010) & 0.000 (0.000) & 0.800 (0.017) & 0.144 (0.060) \\
0.9 & 0.932 (0.010) & 0.000 (0.000) & 0.932 (0.010) & 0.000 (0.000) & 0.787 (0.015) & 0.046 (0.027) \\
1.0 & 0.932 (0.010) & 0.000 (0.000) & 0.932 (0.010) & 0.000 (0.000) & 0.777 (0.015) & 0.000 (0.000) \\
\hline
\end{tabular}%
}
\end{table}

In this section, we describe the experimental evaluation. All experiments are implemented in Python, using the scikit-learn~\cite{tdp_scikit-learn} and DESLib~\cite{tdp_deslib} libraries. We use simple and balanced datasets: Digits~\cite{tdp_digits}, Vehicle~\cite{tdp_vehicle}, and Wine~\cite{tdp_wine}. The models are evaluated using 5-times repeated stratified 2-fold cross-validation. %The results of each cross-validation are averaged. 
A crucial aspect of DS is the generation of a pool of classifiers. A common practice in the literature on DS is to use the Bagging method to generate a pool containing base classifiers that are both diverse and informative. We use a pool of classifiers generated using the Random Forest method. %Random Forest is a state-of-the-art algorithm that combines the concepts of bagging and random subspace. 
The pool of base classifiers consists of an RF containing 25 estimators with a maximum depth of 10. The \textit{K}-means algorithm is used to define the competence region (with \textit{k}=5). In this experiment, we vary the IH threshold and measure the performance of DES-Clustering, KNORA-U, and KNORA-E. Neighborhood-based methods are used only for comparison, as they are not suited for TinyML systems. When a test instance, depending on the IH threshold, is considered easy to classify, it will be classified by a pruned decision tree (for DES-Clustering) or kNN (for KNORA-U and KNORA-E). Following preliminary experiments, the maximum depth of the pruned decision tree is 10. The choice of this value is crucial, as it affects not only accuracy but also energy consumption. We report the accuracy metric and the energy proxy, which is an approximate measure of energy consumption calculated as a percentage of samples sent to the heavier DES classifier. %The bolded values in the tables indicate the threshold values for which a given configuration lies on the Pareto front (maximizing accuracy, minimizing the energy proxy).
Table~\ref{tab:threshold_results} presents the results of the experimental evaluation. In general, with the exception of KNORA-U for Wine and KNORA-E for Digits, accuracy values show a downward trend for all methods. The energy proxy values are always decreasing and never increasing. The dynamics of the energy proxy are significantly different, where for the KNORA methods, the number of samples classified by the light classifier reaches its maximum faster. In the case of DES-Clustering, the energy proxy values smoothly transition from 1.0 to 0.0 for the Vehicle and Digits datasets. For the Wine dataset, we observe a point at which there is a jump in both the energy proxy and accuracy. The results indicate that appropriate threshold selection can significantly reduce the energy cost with a limited decrease in classification quality. 

\section{Conclusions}\label{sec:conclusions}
In this paper, we proposed using TDP as an IH metric in combination with the DES-Clustering method as a memory- and time-efficient solution tailored to TinyML systems. The results indicate that selecting an appropriate threshold can significantly reduce energy consumption with only a limited decrease in classification quality. This method allows us to adjust classification accuracy, thereby influencing computational complexity and energy consumption for inference. The presented approach can potentially be used in any TinyML application where balancing accuracy and energy consumption is critical. In future, we plan to extend the concept by controlling the threshold based on context (e.g., battery level or photovoltaic panels energy availability). This will enable the system to charge the battery and dynamically adjust the threshold value in real time. %Further work may also include exploring additional use cases.

%We presented preliminary findings on a novel application of the TDP IH method to the TinyML system. We conducted demonstration experiments to show the behavior of the proposed approach in a controlled environment. 

\bibliographystyle{splncs04}
\bibliography{mybiblo}

@article{tdp_1a,
  author  = {Britto, Alceu S. and Sabourin, Robert and Oliveira, Luiz E. S.},
  title   = {Dynamic Selection of Classifiers---A Comprehensive Review},
  journal = {Pattern Recognition},
  volume  = {},
  number  = {},
  pages   = {},
  year    = {2014}
}

@article{tdp_2a,
  author  = {Cruz, R. M. O. and Sabourin, R. and Cavalcanti, G. D.},
  title   = {Dynamic Classifier Selection: Recent Advances and Perspectives},
  journal = {Information Fusion},
  volume  = {},
  pages   = {},
  year    = {2018}
}

@inproceedings{tdp_6a,
  author    = {Soares, R. G. and others},
  title     = {Using Accuracy and More Diverse to Select Classifiers to Build Ensembles},
  booktitle = {International Joint Conference on Neural Networks},
  year      = {2006},
  pages = {}
}

@article{tdp_tinySurvex,
  author  = {Abadade, E.},
  title   = {A Comprehensive Survey on {TinyML}},
  journal = {IEEE Access},
  volume  = {},
  pages   = {},
  year    = {2023}
}

@article{tdp_ensemble,
  author  = {Mienye, I. and Sun, Y.},
  title   = {A Survey of Ensemble Learning: Concepts, Algorithms, Applications, and Prospects},
  journal = {IEEE Access},
  volume  = {},
  pages   = {},
  year    = {2022}
}

@article{tdp_scikit-learn,
  author  = {Pedregosa, F. and others},
  title   = {Scikit-learn: Machine Learning in Python},
  journal = {JoMLR},
  volume  = {},
  pages   = {},
  year    = {2011}
}

@INPROCEEDINGS{tdp_bigLITTLE,
  author={Park, Eunhyeok and others},
  booktitle={2015 CODES+ISSS}, 
  title={Big/little deep neural network for ultra low power inference}, 
  year={2015},
  volume={},
  number={},
  pages={},
  }

@INPROCEEDINGS{tdp_paper0,
  author={Cruz, Rafael M. O. and others},
  booktitle={2017 IPTA}, 
  title={Dynamic ensemble selection VS K-NN: Why and when dynamic selection obtains higher classification performance?}, 
  year={2017},
  volume={},
  number={},
  pages={},
  doi={}}

@INPROCEEDINGS{tdp_paper1,
  author={Dantas, Carine and others},
  booktitle={2019 BRACIS}, 
  title={Instance Hardness as a Decision Criterion on Dynamic Ensemble Structure}, 
  year={2019},
  volume={},
  number={},
  pages={},
  doi={}}

@misc{tdp_wine,
  author       = {Aeberhard, Stefan and Forina, M.},
  title        = {{Wine}},
  year         = {1992},
}

@misc{tdp_vehicle,
  author       = {Mowforth, Pete and Shepherd, Barry},
  title        = {{Statlog (Vehicle Silhouettes)}},
  note         = {},
    year         = {1987},
}

@misc{tdp_digits,
  author       = {Alpaydin, E. and Kaynak, C.},
  title        = {{Optical Recognition of Handwritten Digits}},
  year         = {1998},
  note         = {}
}

@article{tdp_deslib,
    author  = {Rafael M. O. Cruz and others},
    title   = {DESlib: A Dynamic ensemble selection library in Python},
    journal = {Journal of Machine Learning Research},
    year    = {2020},
    volume  = {},
    number  = {},
    pages   = {},
    url     = {}
}

@article{tdp_warden,
author = {Warden, Pete and others},
title = {Machine Learning Sensors},
year = {2023},
issue_date = {November 2023},
publisher = {Association for Computing Machinery},
address = {},
volume = {},
number = {},
issn = {},
journal = {Commun. ACM},
pages = {},
numpages = {}
}

@article{tdp_LorenaTrusting,
author = {Lorena, Ana C. and others},
title = {Trusting My Predictions: On the Value of Instance-Level Analysis},
year = {2024},
issue_date = {July 2024},
publisher = {Association for Computing Machinery},
address = {New York, NY, USA},
volume = {},
number = {},
issn = {0360-0300},
journal = {ACM Comput. Surv.},
articleno = {},
numpages = {},
}

@article{tdp_Smith2014,
author = {Smith, Michael R. and others},
title = {An instance level analysis of data complexity},
year = {2014},
issue_date = {May       2014},
publisher = {Kluwer Academic Publishers},
address = {USA},
volume = {},
number = {},
issn = {0885-6125},
journal = {ML},
pages = {},
numpages = {},
}

@BOOK{tdp_Breiman,
  title     = "Classification And Regression Trees",
  author    = "Breiman, Leo and others",
  publisher = "Routledge",
  year      =  1984
}

@article{tdp_tinySurvey,
author = {Capogrosso, Luigi and others},
year = {2024},
month = {},
pages = {},
title = {A Machine Learning-Oriented Survey on Tiny Machine Learning},
volume = {},
journal = {IEEE Access},
}

@InProceedings{tdp_item24,
author="Pu{\'{s}}lecki, Tobiasz and Walkowiak, Krzysztof",
title="On The Dynamic Ensemble Selection for TinyML-based Systems - a Preliminary Study",
booktitle="ITEM ECML-PKDD 2024",
year="2026",
}

@InProceedings{tdp_ITEM25,
author="Puslecki, Tobiasz and Walkowiak, Krzysztof",
title="Data Stream Processing for Resource-Constrained TinyML Systems",
booktitle="ITEM ECML-PKDD 2025",
year="2026",
}

\end{document}